\documentclass[11pt,a4paper]{article}

\usepackage{times,latexsym}
\usepackage{url}
\usepackage{amsthm}
\usepackage{amsmath}
\usepackage[utf8]{inputenc}
\usepackage{todonotes}
\usepackage[english]{babel}
\usepackage[style=american]{csquotes}
\usepackage[acceptedWithA]{tacl2018v2}
\usepackage{subcaption}
\usepackage{graphicx}
\usepackage{tabularx}
\usepackage{xcolor}
\usepackage{xspace,mfirstuc,tabulary}

\newif\iftaclinstructions
\taclinstructionsfalse 
\iftaclinstructions
\newcolumntype{H}{>{\setbox0=\hbox\bgroup}c<{\egroup}@{}}

\renewcommand{\confidential}{}
\renewcommand{\anonsubtext}{(No author info supplied here, for consistency with
TACL-submission anonymization requirements)}
\newcommand{\instr}
\fi

\iftaclpubformat 

\else

\fi

\newcommand{\wordmetric}{\emph{Word-Count Metric}}
\newcommand{\usermetric}{\emph{User-Count Metric}}

\newcommand{\igrword}{\emph{IGR-Words}}
\newcommand{\igruser}{\emph{IGR-Users}}

\newcommand{\wordrank}[1]{\emph{Word-Count%
    \ifthenelse{\equal{#1}{*}}{}{ Ranking}%
}}
\newcommand{\userrank}[1]{\emph{User-Count%
    \ifthenelse{\equal{#1}{*}}{}{ Ranking}%
}}

\DeclareMathOperator*{\argmax}{arg\,max}  

\title{Exploiting user-frequency information for mining regionalisms\\ from Social Media texts}

\author{
    Juan Manuel Pérez \\ Universidad de Buenos Aires, CONICET \\ jmperez@dc.uba.ar 
    \And
    Damián E. Aleman \\  Universidad de Buenos Aires \\ \\ daleman@dc.uba.ar
    \AND
    Santiago N. Kalinowski \\ Academia Argentina de Letras \\ s.kalinowski@aal.edu.ar
    \And
    Agustín Gravano \\ Universidad de Buenos Aires, CONICET \\ gravano@dc.uba.ar
}
\date{}

\begin{document}
\maketitle
\begin{abstract}
The task of detecting regionalisms (expressions or words used in certain regions) has traditionally relied on the use of questionnaires and surveys, and has also heavily depended on the expertise and intuition of the surveyor.
The irruption of Social Media and its microblogging services has produced an unprecedented wealth of content, mainly informal text generated by users, opening new opportunities for linguists to extend their studies of language variation.
Previous work on automatic detection of regionalisms depended mostly on word frequencies. In this work, we present a novel metric based on Information Theory that incorporates user frequency.
We tested this metric on a corpus of Argentinian Spanish tweets in two ways: via manual annotation of the relevance of the retrieved terms, and also as a feature selection method for geolocation of users.
In either case, our metric outperformed other techniques based solely in word frequency, suggesting that measuring the amount of users that produce a word is informative.
This tool has helped lexicographers discover several unregistered words of Argentinian Spanish, as well as different meanings assigned to registered words.

\end{abstract}

\section{Introduction}

Lexicography is the art of writing (designing, compiling, editing) dictionaries: that is, the description of the vocabulary used by members of a speech community \cite{atkins2008oxford}. In the last 30 years, tools coming from Computational Linguistics have helped with this kind of work, mainly in the form of corpora of selected texts. Statistical analysis of corpora results in evidence to support the addition or removal of a word from a dictionary, its marking as dated or unused, as regional, etc., depending on different criteria.

In the process of compiling dictionaries, differences emerge between dialects, where frequently certain words or meanings do not span across all speakers. Since languages are ideal constructs based on the observation of dialects, it is of paramount importance to establish which words are most likely to be shared by an entire linguistic community and which are only used by a smaller group. In this last case, the description profits greatly from information as precise as possible, about geographical extension (region, province, district, city, even neighborhood), about registry (colloquial, neutral, formal), about frequency (actual, past or a combination of both depending on chronological span of the corpus), or any other variable.

Words that are used exclusively or mainly in a particular subregion of the territory occupied by a linguistic community, or that are used there with a different meaning, are called \emph{regionalisms}, localisms or dialectal words. For example, the words ``che''\footnote{Interjection used to get the interlocutor's attention.} and ``metegol''\footnote{Mechanic game that emulates football (\textit{futbolín}) \cite{academia2008diccionario}.} are used more frequently in Argentina than in Spain. Such words are commonly detected through surveys \cite{almeida1995variacion, labov2005atlas} or transcriptions, using methods that depend more or less on the intuition and expertise of linguists. The results of this methodology are of great value to lexicographers, who need evidence to support either the addition of a word into a regional dictionary or the indication of where it is used. Information gathered with these traditional methods has been used as lexical variables to computationally calculate similarities in dialects \cite{kessler1995computational, nerbonne1996phonetic}.

The irruption of Social Media and its microblogging services has produced an unprecedented wealth of content, with a clear tendency towards informal or colloquial text generated by users. This opens many opportunities to linguists due to the possibility of accessing geotagged contents, which provide valuable information about the origin of users. Social media texts have been used to study dialects and establish ``continuous'' isoglosses \cite{gonccalves2014crowdsourcing, huang2016understanding}, to study language diffusion \cite{eisenstein2014diffusion} and other linguistic studies. 


A problem intimately related to lexical dialectology is that of \emph{geolocation}. These can be seen as inverse problems: one maps regions into dialectal words; the other maps words to regions (locations) \cite{eisenstein2014identifying}. Thus, a way to assess dialectometric models is to use them in geolocation algorithms. In fact, regionalisms can be seen as \emph{location-indicative words} \cite{han2012geolocation}.

Most previous work in word-centric geolocation algorithms (and lexical dialectology) relies on the observation of the frequency of a certain word, ignoring the number of users producing them. Also, very little work has been performed in Spanish.

In this work, we present an information-theoretic measure to detect regionalisms in Social Media Texts, particularly on Twitter, and we test it against a dataset of tweets in Argentinian Spanish. Our contributions are twofold: a) we introduce a new metric based on Information Theory which can be seen as a mixture of \emph{TF-IDF} and Information Gain; and b) we show that measuring the dispersion of users is a strong indicator of relevance, for both lexical dialectology and geolocation.
We conduct our experiments on a dataset of tweets in Argentinian Spanish, with 81M tweets, 56K users, all balanced across the country's 23 provinces.

\section{Previous Work}

Most of the previous work in lexical dialectometry consists in measuring words known a priori to be regional variants. 
These works typically use features gathered from sources such as web searches \cite{grieve2013site} and manually-collected regionalisms \cite{ueda2003varilex, kessler1995computational}. Even works analyzing data from Twitter \cite{huang2016understanding, gonccalves2014crowdsourcing} still rely on words known a-priori to discover dialectal patterns.

Language evolves so quickly that it is important to detect these contrastive words automatically --or at least, moderate the efforts to detect them. Two types of approaches exist for this problem: one model-based and one metric-based \cite{rahimi2017continuous}.

Model-based approaches use generative models to detect topics and regional variants \cite{eisenstein2010latent, ahmed2013hierarchical}. Topic modelling such as these approaches suffer from being very algorithmically complex, thus limiting the amount of data they can process.

Metric-based approaches \cite{cook2014statistical, chang2012phillies, jimenez2018automatic, monroe2008fightin} create a statistic for each word or expression, and then rankings of each expression. The generated lists of words could be evaluated by checking an external source of regionalisms --such as a thesaurus or dictionary. These methods are usually faster and more scalable but might get corrupted by topics.

In particular, we compare our metrics with those of \citet{han2012geolocation}: Term-Frequency Inverse Location Frequency (TF-ILF) and Information-Gain Ratio. We refer to them in the following section.

Text-based geolocation can be seen as the inverse problem of dialectology: while dialectology maps regions to text, geolocation maps text to regions \cite{eisenstein2014identifying}. Thus, a reasonable way of assessing the performance of a method for discovering regional words is to use this as feature-selection method for a geolocation classifier, as performed in \citet{han2012geolocation}. In this work, we used provinces as our unit of study, but finer grained geolocation could be performed by using an adaptive grid  \cite{roller2012supervised}.

\citet{rahimi2017neural} proposes a different approach to this problem: the authors train a multilayer perceptron with bag-of-words as input to geolocate users. Intermediate layers serve as vector representations to perform lexical analysis by analyzing proximity in the embedding space.

Information Theory is one of the basis of many of these methods \cite{han2012geolocation, roller2012supervised, chang2012phillies}. Other uses of information-theoretic measures include telling whether a hashtag is promoted by spammers by analyzing its dispersion in time and users \cite{Cui:2012:DBE:2396761.2398519, ghosh2011entropy}, and also to discover valuable features from users messages on Twitter for sentiment analysis and opinion mining \cite{pak2010twitter}. The metrics in the next section use this concept of measuring the entropy of the users of a word.

\section{Method and Materials}

\subsection*{Data}
\begin{figure*}[h]
    \centering
    \includegraphics[width=\textwidth]{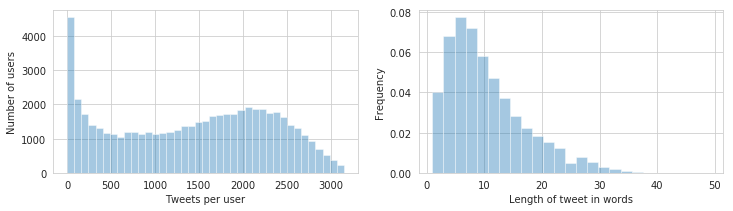}
    \caption{Distributional figures of the dataset. Left: Distribution of number of tweets per user. Right: Distribution of length (in words) of tweets.}
    \label{fig:tweets_distribution}
 \end{figure*}

To gather our data, information of \textit{departments} in Argentina (the second-level administrative division of the country, after \textit{provinces}) was collected from the 2010 National Census.\footnote{\url{https://www.indec.gov.ar}} Next, a lookup was made through the Twitter API for users with location matching those departments.

Although location fields in Twitter are not to be trusted most of the times \cite{hecht2011tweets} as we restrict it to a fixed number of names (departament names) most of the noise is reduced. The Python library \textit{tweepy} was used to interact with the Twitter API.

For each of these users, we retrieved their entire tweetlines. Tweets were tokenized using \emph{NLTK} \cite{bird2009natural}. Hashtags and mentions to users were removed; the remaining words were downcased; and identical consecutive vowels were normalized up to three repetitions (``woaaa'' instead of ``woaaaaaa'').
Table \ref{tab:summary_tweets} lists the figures for the collected dataset, and Figure \ref{fig:tweets_distribution} display the distributions of tweets per user and length of tweets.

\begin{table}[]
\begin{center}

\begin{tabular}{lrrr}
            & Total   & Mean   & SD \\ 
\hline
Words       &  647M   &  28.14M & 6.64M  \\ 
Tweets      &  80.9M  &  3.51M  & 0.91M  \\ 
Users       &  56.2K  &  2.44K  & 0.04K  \\ 
Vocabulary  &   7.5M  &  0.32M  & 0.04M  \\ 
\hline
\end{tabular}

\caption{Dataset summary. Total figures are provided, along with province-level mean and standard deviation.}
\label{tab:summary_tweets}
\end{center}
\end{table}

It is well known that Twitter vocabulary tends to be very noisy \cite{kaufmann2010syntactic} with lots of contractions, non-normal spellings (e.g., vocalizations), typos, etc. Consequently, only words occurring more than 40 times and used by more than 25 users were taken into account. This removes about 1\% of the total words and reduces vocabulary from 2.3 million words to around 135 thousand words.

\subsection*{Method}
We can think of a \emph{regionalism} as a word whose usage is not uniform across all the studied territory -- i.e., whose concentration is high in a specific region of the country. We are trying, in fact, to measure the \textit{disorder} in the usage of a word, and there exists a specific information-theoretic tool for this: entropy.

It is known that entropy holds information about the semantic role played by a word. Given a text, high-entropy words are more likely to be pronouns, connectors and other closed-class words, whereas its low-entropy counterparts are usually nouns and adjectives with fuller semantic content \cite{montemurro2002entropic, montemurro2010towards}.

Taking into account their number of occurrences, words with high entropy (i.e., high disorder) can be regarded as used evenly all across the country. On the other hand, low-entropy words are used with higher frequency in a few specific locations.

Let $l_1, l_2, \ldots l_N$ be our locations, and $\omega_1,$ $\omega_2,$ $\ldots \omega_M$ our vocabulary. If $O_j$ refers to the event of occurrence of word $\omega_j$, then $p(l_i | O_j)$ denotes the probability that word $w_j$ occurred in location $l_i$.

We next define the \emph{word-count entropy} as
\begin{equation}
    H_\text{words}(\omega_j) = -\sum \limits_{i=1}^{N} p(l_i| O_j) \cdot \log p(l_i| O_j)
    \label{eqn:Hw}
\end{equation}

Note that this measure does not take into account the actual frequency of words. For instance, if two words $\omega_1$ and $\omega_2$ occur only in one particular location, but $\omega_1$ is much more frequent than $\omega_2$, both words will still have the same entropy according to Equation \ref{eqn:Hw}.

In a similar fashion to \emph{tf-idf} and inspired by \citet{montemurro2010towards} and \citet{han2012geolocation}, we define measure $I_\text{words}(\omega)$ for word $\omega$ as follows:
\begin{equation}
  \label{eqn:iv_words}
  I_\text{words}(\omega) = p(\omega) \cdot (\log N  - H_\text{words}(\omega)),
\end{equation}
where $\log N$ is the maximum possible value of $H_\text{words}(\omega)$ \cite{shannon2001mathematical}, and $p(\omega)$ is the relative frequency of $\omega$ in the corpus ($0 \leq p(\omega) \leq 1$). In this way, $I_\text{words}(\omega)$ will be high for frequent words that accumulate in just a few locations.

Another important aspect of a word is the amount of people using it on Twitter \cite{Cui:2012:DBE:2396761.2398519}. Assuming we are now sampling users, let $U_j$ be the event that a particular user uses word $\omega_j$. Then $p(l_i| U_j)$ denotes the probability that the location of a user is $l_i$ given the fact that s/he uses word $\omega_j$. 
We define the \emph{user-count entropy} as

\begin{equation}
    H_\text{users}(\omega_j) = -\sum \limits_{i=1}^{N} p(l_i| U_j) \cdot \log p(l_i| U_j)
\end{equation}

\noindent and the following metric of $\omega$,
\begin{equation}
  \label{eqn:iv_users}
  I_\text{users}(\omega) = q(\omega) \cdot (\log N - H_\text{users}(\omega)),
\end{equation}
where $q(\omega)$ is the proportion of users who mentioned $\omega$ in the corpus ($0 \leq q(\omega) \leq 1$). Note that $I_\text{users}(\omega)$ will be high for words mentioned by several users who accumulate in just a few locations.


According to Zipf's Law, the frequencies of top-used words are many orders of magnitude higher than others -- a phenomenon also true when counting users of words. So the $p(\omega)$ and $q(\omega)$ terms in equations \eqref{eqn:iv_words} and \eqref{eqn:iv_users} become a problem as words with high frequencies overcome their low entropies. To alleviate this, we performed a normalization on the word frequency as follows. Let $M_\omega$ be the most-frequent word, that is,
\begin{equation}
    M_\omega = \argmax\limits_{\omega \in W} \#\omega,
\end{equation}
where $\#\omega$ denotes the total number of occurrences of $\omega$ in our dataset. Then, the \emph{Normalized log-frequency} of word occurrences is defined as
\begin{equation}
    n_\text{words}(\omega) = \frac{\log(\# \omega)}{\log(\# M_w)}
\end{equation}

Words with very high frequency differ little on their values of $n_\text{words}(\omega)$. We define analogously the \emph{Normalized log-frequency} of user mentions $n_{users}$. Hence, we redefine our two metrics as
\begin{align}
  I_\text{words}(\omega) &= n_\text{words}(\omega) (\log(n) - H_\text{words}(\omega)) \\
  I_\text{users}(\omega) &= n_\text{users}(\omega) (\log(n) - H_\text{users}(\omega))
\end{align}

\noindent
We call the first metric \emph{Log-Term Frequency Information Gain (LTF-IG)} and the second one \emph{Log-User Frequency Information Gain (LUF-IG)}.

A word having a high value for the metrics just defined may be regarded as being more present in a certain region than in the rest of the country. We subsequently sort all words in our dataset relative to these metrics, thus obtaining two word rankings: \wordrank{} and \userrank{}. The words that appear in the first positions of a ranking are those with high values for the metric, and thus more likely to be regionalisms.





\subsection{Lexicographic Validation}


With these rankings, a team of lexicographers 
performed a linguistic validation of the first thousand words according to each metric. This qualitative analysis consisted in a detailed study, word by word, to determine if the word in question is part of the lexical repertoire of a community of speakers. Proper and place names (toponyms) were excluded --as is traditional in lexicography-- although many words in this class had high values for our metrics.
To facilitate the exclusion of regionalisms by lexicographers, words suspected of being toponyms were automatically highlighted.

To perform the linguistic validation, lexicographers were provided with tables containing figures for each word and province: number of users, number of occurrences and normalized frequency (occurrences per million words). Also, samples of tweets containing these words were provided when necessary.

As a result of this process, every word in the top-1000 of each ranking was annotated with `1' if it had lexical relevance as a regionalism, or `0' if it had not. Lastly, lexicographers performed a characterization of the words marked as regionalisms, according to the linguistic phenomenon they represent. The outcome of these procedures is described in the following sections.

\subsection{Feature Selection Methods for Geolocation}

To indirectly assess the pertinence of our metrics, we used each as a feature-selection method to train geolocation classifiers. 
This means that, instead of using the entire bag-of-words as input for a geolocation algorithm, we consider a smaller subset of the vocabulary. This dimensionality reduction of the feature space is aimed at boosting the classifier performance.

This approach to geolocation can be classified as ``word-centric'', as it uses lexical information from tweets to predict a location \cite{zheng2018survey}. We are concerned with \emph{user} geolocation -- i.e., not tweet geolocation. Thus, the units or documents considered are all the tweets from single users. From the collected dataset, we randomly selected 10,000 users, with 7,500 used as training set and 2,500 for testing purposes.

For reference, we compare our results to those obtained using the \emph{Information Gain Ratio (IGR)} metric as described in \citet{han2012geolocation, cook2014statistical}: if $L$ is a random variable denoting the location of a given occurrence of a $\omega_i$, then the \emph{Information Gain} of $\omega_i$ is
\begin{align*}
    IG(\omega_i) =& H(L) - H(L|\omega_i)\\
                \propto& P(\omega_i) \sum\limits_{j=1}^{m} P(c_j | w_i) \log P(c_j| w_i)\\
                +& P(\overline{w_i}) \sum\limits_{j=1}^{m} P(c_j | \overline{w_i}) \log P(c_j| \overline{w_i})
\end{align*}
where $P(\overline{\omega_i})$ denotes the probability that $\omega_i$ does not occur. Then, $IGR(\omega_i)$ is defined as

\begin{equation}
    IGR(\omega_i) = \frac{IG(\omega_i)}{IV(\omega_i)}
\end{equation}
where $IG$ is normalized by 
\begin{equation*}
    IV(\omega) = - P(\omega) \log P(\omega) - P(\overline{\omega}) \log P(\overline{\omega}))
\end{equation*}

We also calculate $IGR$ but with user-frequencies, in a similar way to Equation \ref{eqn:iv_users}. As a baseline for our feature selection methods, we also calculate \emph{Term-Frequency Inverse Location Frequency (TF-ILF)}, which consists in sorting our terms first by Location Frequency (in ascending order) and then by Term-Frequency (in descending order).

Summing up, five feature selection methods are tested as feature selection for geolocation: \emph{TF-ILF}, \emph{LTF-IG}, \emph{LUF-IG}, basic \emph{IGR}, and \emph{User IGR}. We train Multinomial Logistic Regressions using the top $N\%$ words as features, and test against the 2.5K held out users. Performance is assessed using accuracy and mean distance between capital cities of each province -- a fairly good estimation, since most of the population concentrates around those cities.

\section{Results}
\begin{table}[t]
  \centering
    \begin{tabular}{lrr}
    {Rank} &       Word &        User     \\
    \hline
    1  &        ushuaia &          chivil \\
    2  &          rioja &            ush  \\
    3  &      chivilcoy &            poec \\
    4  &        bragado &\textbf{malpegue}\\
    5  &         viedma &   \textbf{aijue}\\
    6  &        logroño &         tolhuin \\
    7  &         chepes &        vallerga \\
    8  &          oberá &  \textbf{yarca} \\
    9  &  \textbf{cldo} &             blv \\
    10 &            tdf &          portho \\
    11 &       riojanos &          jumeal \\
    12 &         breñas &   \textbf{sinf} \\
    13 &         choele &        plottier \\
    14 &       gallegos &           kraka \\
    15 &      tiemposur &             fsa \\
    16 &      fueguinos & \textbf{bombola}\\
    17 &      chilecito &  \textbf{yarco} \\
    18 &            blv &       sanagasta \\
    19 &            ush &            wika \\
    20 &          merlo &           obera \\
    \hline
  \end{tabular}
\caption{Top 20 words for the two metrics. Words in bold have lexicographic interest as regionalisms.}
\label{tab:20_top_words}
\end{table}

Table \ref{tab:20_top_words} shows the top-20 words calculated with each metric.  Many are toponyms: \emph{chivil, ush, blv, tolhuin, kraka, sanagasta, wika} refer to towns, cities and local clubs. Also, some words refer to gentilics (\emph{riojanos, fueguinos}), or local institutions (\emph{POEC}). Some of these words emerge as regionalisms: \emph{yarca/yarco, aijue, sinf, cldo, bombola, malpegue}. We can observe that many words are shared among the rankings. \userrank*{} and \wordrank*{} have an overlap of 63\% in the top thousand words.

\begin{figure*}[t]
   \centering
   \begin{subfigure}[t]{0.49\textwidth}
   \includegraphics[width=\textwidth]{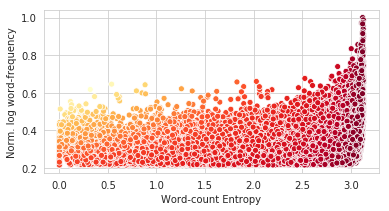}
   \caption{Color scale: \wordrank{}}
   \label{fig:word_iv_word_axes}
   \end{subfigure}
   \begin{subfigure}[t]{0.49\textwidth}
   \includegraphics[width=\textwidth]{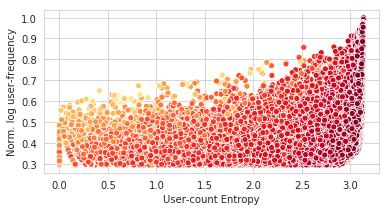}
   \caption{Color scale: \wordrank{}}
   \label{fig:word_iv_user_axes}
   \end{subfigure}
   \begin{subfigure}[t]{0.49\textwidth}
   \includegraphics[width=\textwidth]{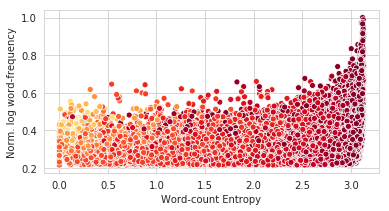}
   \caption{Color scale: \userrank{}}
   \label{fig:user_iv_word_axes}
   \end{subfigure}
   \begin{subfigure}[t]{0.49\textwidth}
   \includegraphics[width=\textwidth]{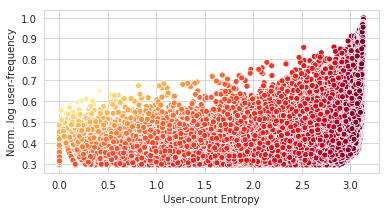}
   \caption{Color scale: \userrank{}}
   \label{fig:user_iv_user_axes}
   \end{subfigure}

   \caption{Scatter plots showing words (dots) along three dimensions. Horizontal axes: word-count entropy $H_\text{words}$ (left plots) or user-count entropy $H_\text{users}$ (right plots). Vertical axes: normalized log word frequencies $n_\text{words}$ (left plots) or user frequencies $n_\text{users}$ (right plots). Color: log word rank according to \wordrank*{} (top plots) or to \userrank*{} (bottom plots); lighter color means higher rank.}
   \label{fig:ivalue}
\end{figure*}

Figure \ref{fig:ivalue} shows four three-dimensional scatter plots. A dot in these plots corresponds to an individual word in our corpus, and is placed along the horizontal axes according to its word- or user-count entropy ($H_\text{words}(\omega)$ and $H_\text{users}(\omega)$, respectively). Along the vertical axes, each dot is located following its corresponding word or user frequency ($n_\text{words}(\omega)$ and $n_\text{users}(\omega)$). Additionally, each dot is colored according to the position of the word in one of our rankings using a chromatic scale, such that the lighter the dot, the higher the word's rank. For clearer visualization, word rankings are also shown in logarithmic scale.

Figure \ref{fig:word_iv_word_axes} shows that words that figure high in the \wordrank{} (in lighter color) tend to appear closer to the upper-left corner of the plot -- that is, such words are more frequent and their mentions are concentrated in fewer regions. Figure \ref{fig:user_iv_user_axes} shows a very similar thing, now with respect to the number of users that mention the words: words high in the \userrank{} are mentioned by a larger number of users from fewer regions.
These two figures display a gradient from the upper-left corner (words ranked higher, in lighter color) to the lower-right corner (words ranked lower, in darker color).

Figure \ref{fig:word_iv_user_axes} uses horizontal and vertical axes corresponding to users ($H_\text{users}$ and $n_\text{users}$), but colors each word with respect to \wordrank{}. Here we can observe a slight perturbation in the gradient: there are words far from the left-corner that have light colors. From this, we understand that there are words with high \wordrank{} that have low \userrank{}.

Likewise, Figure \ref{fig:user_iv_word_axes} uses \userrank{} to color the points, and word axes $H_\text{user}$ and $n_\text{user}$. The perturbation in the gradient is even clearer in this plot: There are many words that appear high in \wordrank{} (closer to the top-left corner, see Figure  \ref{fig:word_iv_word_axes}) but appear low in \userrank{} (darker color).

To further inspect this phenomenon, we searched for words that have large differences in the logarithm of \wordrank{} and \userrank{}. The logarithm minimizes the difference between words ranked very high (e.g. between the word at position 10,000 and another in position 20,000) and amplifies the difference when one of the ranks is low and the other is high. 
A close examination of these words and the tweets they were used in showed that they were in the vocabulary of bots (news and metheorological accounts, or accounts using applications to get more followers) or small niches of fans of a certain celebrity. From the top-100 words sorted by this difference, only one has a higher ranking in users than in words.

Summing up, when a word has a high \userrank{}, it also tends to have a high \wordrank{}. The reverse is not true, however, as words produced by a small number of accounts would not rank well with respect to users. Thus, the  \userrank{} successfully discards words coming from automatic agents, as already done in \citet{Cui:2012:DBE:2396761.2398519}.

\begin{table}[t]
    \centering

    \begin{tabular}{lrr}
    Word &  Word Rank &  User Rank \\
    \hline
    rioja         &              2 &           2499 \\
    vto           &             27 &          28179 \\
    hoa           &             81 &          83717 \\
    contextos     &             88 &          71290 \\
    cardi         &             32 &          23756 \\
    agraden       &            107 &          75042 \\
    hemmings      &             59 &          40227 \\
    ushuaia       &              1 &            565 \\
    tweeted       &             43 &          21342 \\
    precipitación &             66 &          31042 \\
    \hline
    \end{tabular}

    \caption{Top 10 words with largest difference between their log word rank and their log user rank.}
    \label{tab:table_rank_differences}
\end{table}

The first thousand words in the \wordrank{} were manually analyzed by the lexicographers, who marked 21.9\% as likely regionalisms.
Analogously, from the first thousand words in the \userrank{}, 30.2\% were marked as being lexicographically interesting.
This validation suggests that observing user-frequency dispersion is more relevant when assessing the word as a regionalism.

Lexical characterization is displayed in Table \ref{tab:characterisation}, which displays some groups among the regionalisms found in the analyzed words with examples. A special note is reserved for the group of \emph{Indigenisms}, where a number of words were found coming from \emph{guaraní} (for instance, \emph{mitaí, angá, angaú, nderakore}) and also from \emph{quechua} (\emph{ura}). It is worth mentioning that words coming from \emph{guaraní} —language spoken in Northeastern Argentina, Paraguay, Bolivia and Southwest of Brazil— coincide with the region delimited by \citet{vidal1964espanol}.

\newcommand{\tabinterspace}{\newline\newline}

\begin{table}[ht]
\centering
Colloquialisms
\begin{tabular}{p{0.1\textwidth} p{0.1\textwidth} p{0.2\textwidth}}
\hline
Word & Region & Meaning \\
\hline
culiado & Córdoba & asshole   \\
chombi & Mendoza &  poor in quality\\
carnasas & Neuquén & not classy, inelegant  \\
bolasear & Cuyo & to bullshit \\
aprontar & E. Ríos& to get ready\\
\hline
\end{tabular}
\tabinterspace{}
Indigenisms
\begin{tabular}{p{0.1\textwidth} p{0.1\textwidth} p{0.2\textwidth}}

\hline
ura & Northwest & vagina (quechua)\\
mitaí & Guaranitic & boy \\
angá & Guaranitic & unfortunate \\
\hline
\end{tabular}
\tabinterspace{}
Regional realities
\begin{tabular}{p{0.1\textwidth} p{0.1\textwidth} p{0.2\textwidth}}
\hline
piadinas & San Juan & roll (food) \\
tarefero & Misiones & yerba mate worker \\
POEC & Neuquén & high School exam \\
\hline
\end{tabular}
\tabinterspace{}
Interjections
\begin{tabular}{p{0.1\textwidth} p{0.1\textwidth} p{0.2\textwidth}}

\hline
aijue & Formosa & surprise \\
yirr & Corrientes & joy \\
aiss & Formosa & annoy \\
jiaa & Corrientes & yeehay \\
\hline
\end{tabular}
\tabinterspace{}
Ortographic variations
\begin{tabular}{p{0.1\textwidth} p{0.1\textwidth} p{0.2\textwidth}}

\hline
pesao & Northwest & pesado\\
ql & Northwest & culiado \\
uaso & Córdoba & guaso \\
\hline
\end{tabular}
\tabinterspace{}
Regional Morpheme
\begin{tabular}{p{0.1\textwidth} p{0.1\textwidth} p{0.2\textwidth}}

\hline
raraso & Córdoba & very strange (raro) \\
tardaso & Córdoba & very late (tarde) \\
\hline
\end{tabular}
\caption{Characterization of some of the regionalisms found in the analysis. Each group corresponds to a subjective category found by the lexicographers during the annotation process}

\label{tab:characterisation}
\end{table}

\subsection{Geolocation of users}


\begin{figure*}[t!]
   \centering

   \begin{subfigure}[t]{0.49\textwidth}
   \includegraphics[width=\textwidth]{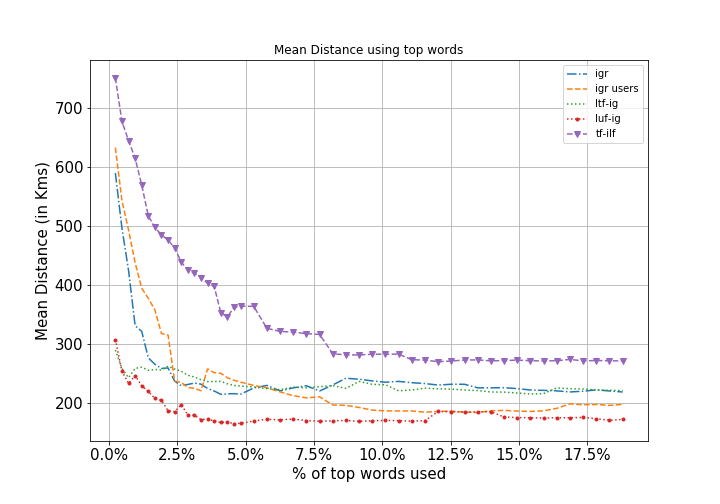}
   \subcaption{Mean distance error in user geolocation}
   \label{fig:mean_distance_comparison}
   \end{subfigure}
   \begin{subfigure}[t]{0.49\textwidth}
   \includegraphics[width=\textwidth]{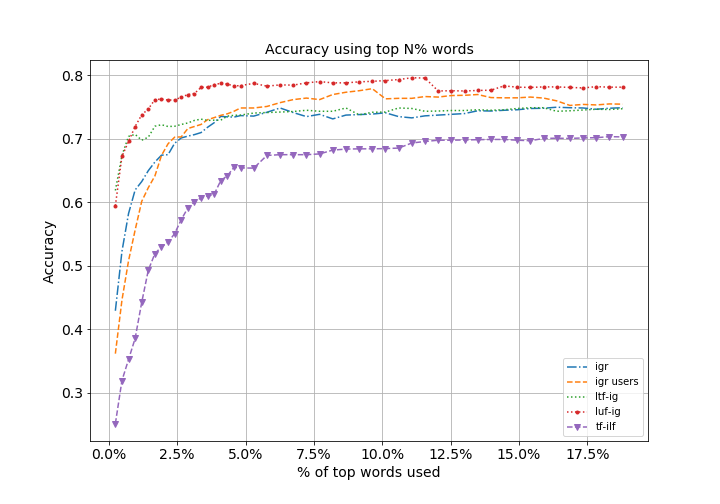}
   \subcaption{Accuracy in user geolocation}
   \label{fig:accuracy_comparison}
   \end{subfigure}

   \caption{Comparison of the metrics when used as feature selection methods for geolocation. Vertical axes show the percentage of the top words used as features to train a Multinomial Logistic Regresion, and vertical axes display the performance of each respective classifier. Figure \subref{fig:mean_distance_comparison} uses mean distance error as y-axis (less is better) and Figure \subref{fig:accuracy_comparison} uses accuracy (more is better)}
   \label{fig:metrics_comparison}
\end{figure*}

\begin{table}[b]
\centering
\begin{tabular}{lrr}
Features         &  Accuracy & Mean Distance  \\
\hline
All           &    0.383 & 599.8 \\
TF-ILF        &    0.654 & 363.3 \\
\igrword{}    &    0.736 & 214.2 \\
\igruser{}    &    0.748 & 234.7 \\
\emph{LTF-IG} &    0.737 & 227.9 \\
\emph{LUF-IG} &    \textbf{0.784} & \textbf{164.9} \\

\hline
\end{tabular}
\caption{Performance of the different feature selection methods when using the top-5000 words.}
\label{tab:geolocation_comparison}
\end{table}

Figure \ref{fig:metrics_comparison} displays the performance of the different feature selection methods when used to train our discriminative classifier. Horizontal axes represent the percentage of top words selected, and the vertical axes represent the accuracy in the case of \ref{fig:accuracy_comparison} and the mean distance error in \ref{fig:mean_distance_comparison}. 

We can observe that comparing both versions of the metrics, those which use user-frequencies obtain better performance than their word-frequency counterparts. This is more clear in the case of \emph{LTF-IG} and \emph{LUF-IG} but we can also observe this in both \emph{IGR} metrics.

\emph{Log User Frequency-Information Gain (LUF-IG)} obtains the best performance geolocating users, and achieves a plateau at about 3.75\%. It outperforms its word-frequency version \emph{LTF-IG} and both IGR metrics. Table \ref{tab:geolocation_comparison} displays the results of using the full bag of words (baseline) versus using the different feature selection methods with 5,000 top words.

\section{Discussion}



Of the proposed metrics, \usermetric{} proved to be the more interesting. It removed from the top positions of the ranking words likely to come from automatic agents or from small niches of users, and lexicographic validation confirmed that this ranking contained more regionalisms than the \wordmetric{}. Further, using this metric as a feature selection method for geolocating users also showed a significative improvement over other metrics -- both its word-frequency counterpart and IGR metrics from \citet{han2012geolocation}.
This might suggest that measuring the dispersion of users of a certain word is a very informative indicator --both in lexicographic and in geolocation terms-- backing what was already found in previous work to detect spam on Twitter \cite{Cui:2012:DBE:2396761.2398519}.


The proposed metric was developed in the context of analyzing regional colloquialisms. This area of the lexicon is most elusive, since its impact on any printed medium arrives noticeably late -- and in many cases it never reaches it at all. Colloquialisms are a class of words hardly found in any other media. Our best performing metric marked as relevant several words that were already listed in the \emph{Diccionario del Habla de los Argentinos} \cite{academia2008diccionario}, a fact that confirms the usefulness of both our metric and Social Media data in general for this task.

An outstanding subgroup found in the analysis are words coming from the \textit{guaranitic} region, in Northeastern Argentina. In particular, three words have been proposed for addition
to the aforementioned dictionary: \emph{angá, angaú, mitaí}. This case is emblematic because it shows how this type of approach can help overcome the intrinsic limitations of doing regional lexicography. When lexicographers are native to only one of the different dialects of the region included in a projected dictionary, the probability of properly detecting and defining words of other dialects is slim or depends on mere chance. As the team of lexicographers expressed when confronted with these three words related to Guaraní heritage, those very robust normalized frequencies across a significant portion of the territory of Argentina would otherwise have remained unknown. Instead of including them in the next edition of the dictionary that attempts to describe all regional lexical items in the country, they would have remained unregistered, thus perpetuating a very serious omission.


As the focus was in detecting lexical variations within provinces, we paid no attention to spatial granularity. If a better granularity were necessary in the analysis, adaptive partitioning could be used \cite{roller2012supervised} to improve geolocation and to find localisms within provinces. Although previous work \cite{vidal1964espanol} indicates that most provinces do not have large dialectal variations within them, this is something that would need to be explored and confirmed in future work.

Also, these techniques should be tested against other datasets (such as those used in \citet{roller2012supervised, han2012geolocation}) to further confirm that they outperform other feature selection methods.  

\section{Conclusions}
In this work, we developed and compared two metrics to detect regionalisms on Twitter based on Information Theory. One was based on the word frequency (\emph{Log Term Frequency-Information Gain, LTF-IG}) and the other on the user frequency of a word (\emph{Log user frequency-Information Gain, LUF-IG}). These metrics may be seen as a mixture of previous information-theoretic measures and classic \emph{TF-IDF}.

We compared their performance by two means. First, a team of lexicographers manually assessed the presence of regionalisms in the first thousand words as ranked by each of these metrics. Second, we tested the metrics as feature-selection methods for geolocation algorithms, for which we also tested against metrics from previous works \cite{han2012geolocation, cook2014statistical}.
In both evaluation types, the metric built upon user frequencies (\emph{LUF-IG}) yielded the better results, suggesting that the number of users of a word is very informative -- perhaps even more than simple word frequency.

This method has aided lexicographers in their task, letting them
propose the addition of a number of words into
the \emph{Diccionario del Habla de los Argentinos}. In the case of this particular dictionary, work relies on a collaborative effort that is based on the intuition of academics and lexicographers that identify regionalisms used mainly (seldom exclusively) within Argentina's borders by carefully parsing over a diversity of sources.
Therefore, using Social Media to automatically detect regionalisms does not limit itself to avoiding most of this manual work, which, in and of itself, would already be a sizeable contribution. Since a considerable portion of the lexical repertoire of a community does not make its way across to published materials (which make most of the 300 millions words included to date in, for example, CORPES XXI \cite{espanolabanco}), the possibility of creating lists of words that are likely to be regional, based on actual utterances written  by users, opens a way of shedding light onto entire pockets of lexical items that would remain otherwise chronically underrepresented in dictionaries. Even when a regional word is published, and then included in corpora, the task of appropriately isolating it remains largely unchanged, given that the word has to previously be identified in order to then take advantage of the statistical information available. 


A further challenge triggered by this work is the detection of regions with different dialectal uses \cite{gonccalves2014crowdsourcing} but using features obtained in a semisupervised fashion with these metrics. This would allow to assess the validity of the dialectal regions of Argentina proposed by Vidal de Battini in 1964 \cite{vidal1964espanol}.
Spatial and temporal information could be also explored, particularly finer-grained locations. Regarding geolocation, the proposed metrics should also be tested against other datasets to evaluate its performance as a feature selection method.

\bibliography{tacl2018}
\bibliographystyle{acl_natbib}

\end{document}